\documentclass[10pt,twocolumn,letterpaper]{article}

\usepackage{wacv}                       

\definecolor{wacvblue}{rgb}{0.21,0.49,0.74}
\usepackage[pagebackref,breaklinks,colorlinks,allcolors=wacvblue]{hyperref}

\def\wacvPaperID{*****}
\def\confName{WACV}
\def\confYear{2027}

\title{\textsc{REDI}: Corpus Aware Patch Ranking for DINOv3 Token Reduction}

\author{
Chanjong Im\\
University of Magdeburg\\
{\tt\small chanjong.im@ovgu.de}
\and
Sebastian Diem\\
University of Hildesheim\\
{\tt\small diem@uni-hildesheim.de}
\and
Thomas Mandl\\
University of Hildesheim\\
{\tt\small mandl@uni-hildesheim.de}
}
\begin{document}
\maketitle
\begin{abstract}
Most token reduction methods for Vision Transformers seek favorable tradeoffs between accuracy and efficiency by pruning, merging, or pooling patch tokens. \textup{\textsc{REDI}} (Relevance for DINOv3 Token Reduction) studies this question through a controlled supervised reference: how should a fixed token budget be allocated across patches for image classification? REDI quantizes final block DINOv3 patch representations into a visual vocabulary and derives class conditioned corpus scores using supervised TF-IDF over visual words. For each validation image, the ground truth class selects a row of the TF-IDF table, and four transformed views produce a TF-IDF map aligned to a reference center crop. A separate dense pass on the same crop provides an attention map. After independent min max normalization, their elementwise product defines the REDI score. A fixed keep, merge, and compress operator then uses score rank to assign patch roles and score magnitude to weight merging and compression.

With precomputed REDI scores, a frozen DINOv3 ViT-B/16 backbone, and the same linear classifier used for dense evaluation, the operator reduces the sequence length from 201 to 107 tokens, a 46.8\% sequence reduction. The REDI variant based on incoming attention mass achieves 84.706\% Top-1 accuracy on ImageNet-1K, compared with 83.514\% for the dense baseline, 82.634\% for incoming attention mass alone, and 81.796\% for supervised TF-IDF alone. The same corpus term also improves reduced classification for three alternative attention formulations relative to their attention only counterparts. Together, these controlled comparisons indicate that class specific corpus statistics and image specific attention provide complementary signals for patch ranking in this setting.
\end{abstract}

\begin{figure*}[t]
\centering
\includegraphics[width=\textwidth]{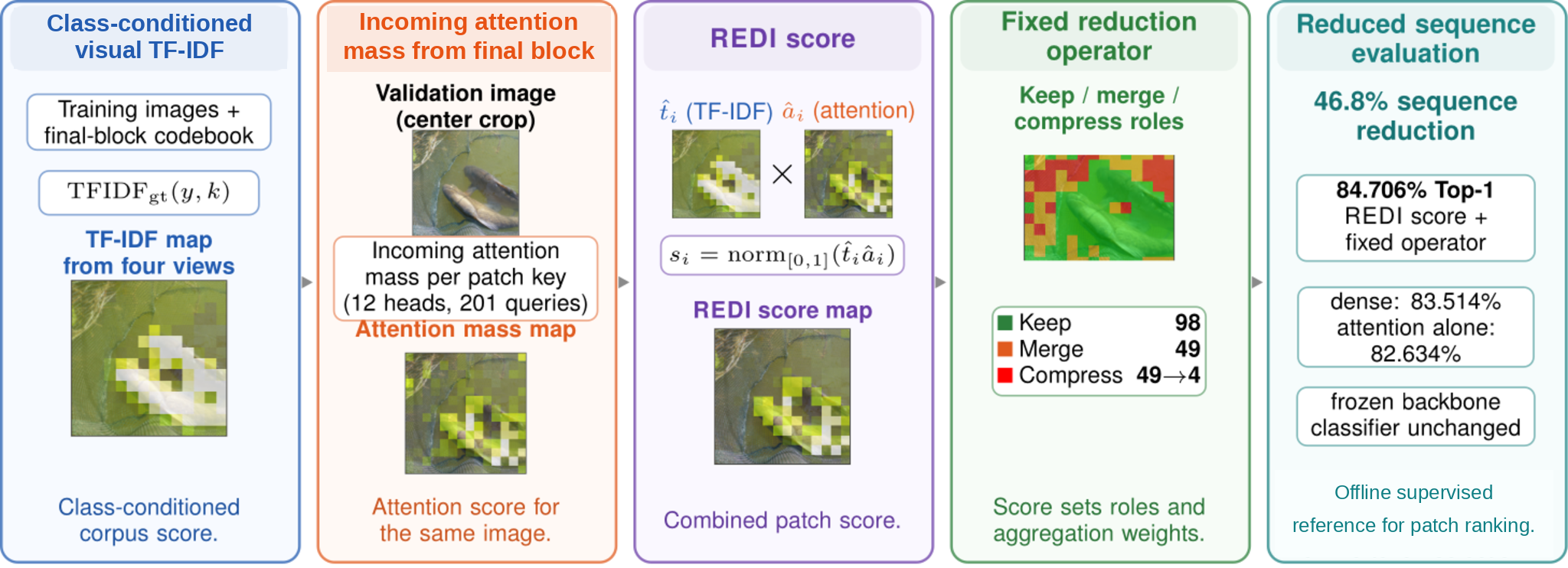}\par
\vspace{0.15em}
\caption{Overview of \textsc{REDI}. Four transformed views provide class-conditioned visual TF-IDF scores aligned to a reference center crop, and a separate dense pass provides incoming attention mass from final-block attention columns. Independent normalization and elementwise multiplication produce the REDI score, which ranks patches for the fixed keep, merge, and compress operator and weights aggregation. With precomputed REDI scores, the frozen backbone and unchanged classifier, the 107 token reduced path reaches 84.706\% Top-1 accuracy.}
\label{fig:overview}
\end{figure*}

\section{Introduction}
Vision Transformers represent an image as a sequence of patch tokens, so sequence length directly affects the cost of self-attention and feedforward layers~\cite{dosovitskiy2021vit}. Token reduction methods exploit this structure by pruning, merging, pooling, or otherwise reorganizing tokens to improve the trade-off between accuracy and computation~\cite{rao2021dynamicvit,liang2022evit,bolya2023tome,marin2023tokenpooling}. This line of work raises a complementary question: how should a limited token budget be allocated across patch representations?

We study this question through a corpus aware relevance formulation. The ground truth class defines the retrieval context, the labeled training images form the collection, and quantized patch representations act as visual words. We construct the vocabulary by applying spherical $k$-means~\cite{dhillon2001spherical} to unit normalized patch representations from the final DINOv3 block and assign each patch to the centroid with the highest cosine similarity. This construction follows the visual word abstraction of classical image retrieval and categorization~\cite{sivic2003videogoogle,csurka2004bags}, while replacing handcrafted local descriptors with frozen DINOv3 features. A visual word receives a high class conditioned corpus score when it is common within the target class, more prevalent there than in the remaining classes, and relatively rare across the training collection.

We introduce \textsc{REDI} (Relevance for DINOv3 Token Reduction), a supervised reference score for class conditioned patch ranking. The score uses the validation image's ground truth class and attention from a dense pass to define a controlled ranking. Its corpus component, $\mathrm{TFIDF}_{\mathrm{gt}}$, is a table indexed by
ImageNet class and visual word. REDI draws on TF-IDF
(term frequency--inverse document frequency), a standard term-weighting framework that balances within document frequency against corpus level rarity~\cite{salton1988term,robertson2004idf}. 

For each validation image, the ground truth class selects a table row, and four transformed views yield a TF-IDF map aligned to the patch grid of a reference center crop. A separate dense pass on that crop yields an attention map. In REDI, the primary image conditioned signal is \emph{incoming attention mass}: we average each patch key column of the final block attention matrix over all heads and query positions. This follows the standard interpretation of self-attention matrices in which rows correspond to query positions and columns correspond to key positions~\cite{guo2023robustifying}; related ViT analyses study how attention behavior varies across supervision regimes~\cite{walmer2023teaching}. REDI independently normalizes the corpus and attention maps, multiplies them elementwise, and normalizes the product to obtain a patch score. Figure~\ref{fig:overview} summarizes the score construction and its use in the reduction operator.

\begin{figure*}[t]
\centering
\includegraphics[width=0.92\textwidth]{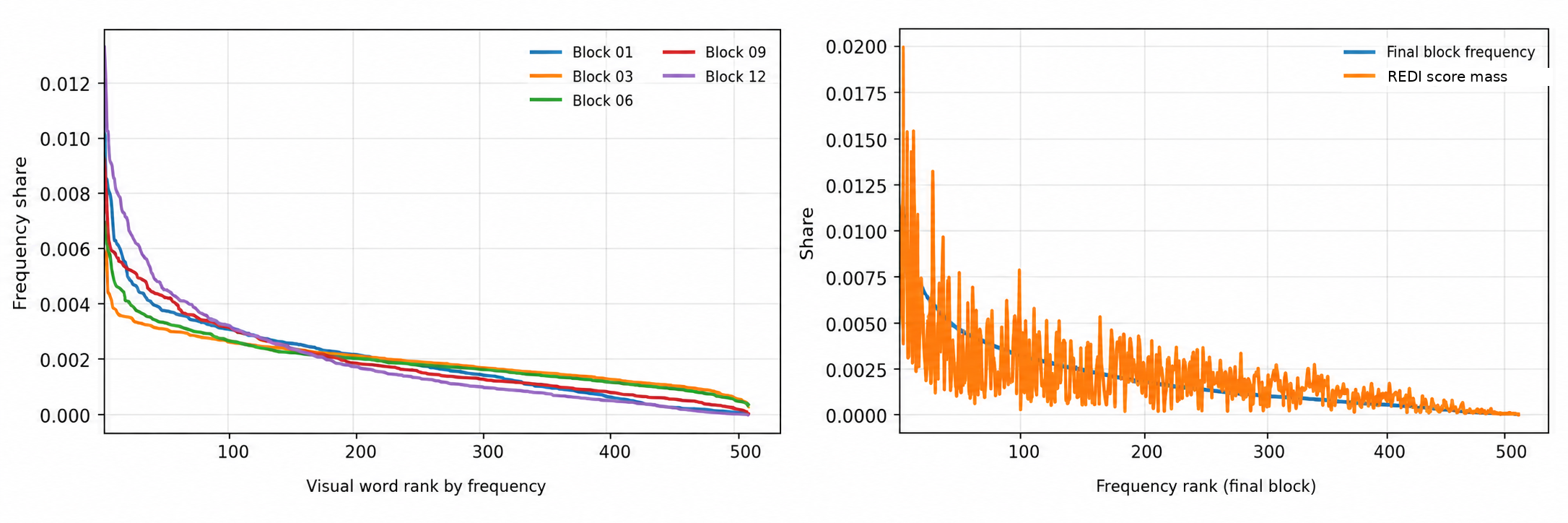}
\caption{Visual word assignment distributions and REDI score mass. Left: visual word frequency by rank for separate $K=512$ analysis codebooks fitted at selected DINOv3 blocks, showing heavy-tailed, Zipf-like usage. Right: final block validation assignments sorted by frequency, with REDI score mass accumulated over the same visual words. The different profiles show that REDI score mass is not determined by raw frequency alone. The analysis codebooks are separate from the codebook used for $\mathrm{TFIDF}_{\mathrm{gt}}$.}
\label{fig:visualvocab}
\end{figure*}

Figure~\ref{fig:visualvocab} characterizes visual vocabulary statistics. Analysis codebooks fitted at selected DINOv3 blocks exhibit heavy-tailed, Zipf-like usage. For the final block, REDI score mass over the same assignments differs from raw assignment frequency, showing that the score is not determined by frequency alone.

We evaluate each scoring signal with the same fixed keep, merge, and compress operator, that is, 98 patch tokens remain explicit, 49 are merged into keep tokens, and 49 are aggregated into four spatial summary tokens. The resulting sequence contains 107 tokens rather than 201. The backbone remains frozen, and the dense feature classifier is reused unchanged, so paired comparisons isolate the scoring signal.

Under this controlled protocol, this pattern holds for all four attention formulations. Every attention only score, as well as $\mathrm{TFIDF}_{\mathrm{gt}}$ alone, falls below the dense Top-1 baseline under the fixed operator, whereas every corresponding REDI product exceeds it. The product with incoming attention mass reaches 84.706\% Top-1 accuracy, compared with 83.514\% for dense evaluation, 82.634\% for incoming attention mass alone, and 81.796\% for $\mathrm{TFIDF}_{\mathrm{gt}}$ alone.

Our contributions are threefold.
\begin{enumerate}[leftmargin=*,nosep]
\item We formulate a class conditioned visual TF-IDF score over a quantized vocabulary of frozen ViT patch features and assign the resulting corpus statistics to individual patches.
\item We combine a class indexed corpus score over quantized ViT patch features with image specific attention to rank individual patches for token reduction. Under the shared DINOv3 reduction protocol, the combination improves reduced classification for all four evaluated attention formulations
relative to their attention only counterparts.
\item We specify a fixed keep, merge, and compress operator for controlled evaluation and report the resulting analytical reductions in sequence length and block level matrix multiplication cost. Once the REDI score is available, the operator reduces sequence length by 46.8\%, the estimated arithmetic cost of the dominant matrix multiplications in each transformer block by 47.8\%, and the quadratic attention term by 71.7\%.
\end{enumerate}

\section{Related work}
\subsection{Self supervised ViT features}
ViT represents an image as a sequence of patch tokens~\cite{dosovitskiy2021vit}. DINO showed that self supervised ViT features exhibit semantically meaningful structure and that attention from the class token maps can reveal object regions without dense supervision~\cite{caron2021dino}. DINOv2 extended this line of work with robust visual features designed to transfer across a broad range of downstream tasks~\cite{oquab2023dinov2}. DINOv3 further scaled self supervised visual pretraining and introduced Gram anchoring to address degradation in dense feature maps during long training schedules~\cite{simeoni2025dinov3}. Register tokens provide dedicated locations for internal computation and reduce high norm artifacts in patch feature maps~\cite{darcet2024registers}.

\subsection{Token pruning and adaptive computation}
Token reduction methods differ in how they identify and process redundant or less informative tokens. DynamicViT inserts lightweight prediction modules at multiple layers and learns progressive, input dependent token sparsification with differentiable attention masking~\cite{rao2021dynamicvit}. A-ViT assigns a halting score to each spatial token and stops updating the token once its accumulated score reaches a threshold~\cite{yin2022avit}. AdaViT adapts patch token, attention head, and transformer block usage to each input~\cite{meng2022adavit}. Evo-ViT separates tokens into informative and placeholder groups using evolved attention from the class token; informative tokens follow the standard transformer path, while placeholder tokens are updated through a representative token at lower cost~\cite{xu2022evovit}.

Several methods propagate token decisions across depth or derive token supervision from later representations or model internal signals. IA-RED$^2$ introduces interpretable modules that progressively remove redundant patches~\cite{pan2021iared2}. Patch Slimming identifies effective patches in the final layer and propagates this supervision to earlier layers~\cite{tang2022patchslimming}. EViT preserves tokens with high attention from the class token and fuses the remaining tokens into a single representative token~\cite{liang2022evit}. ATS is a parameter free differentiable module that scores and adaptively samples tokens for each input~\cite{fayyaz2022ats}. SPViT uses latency aware multi-head token selectors and aggregates pruned tokens into a package token rather than discarding them~\cite{kong2022spvit}. TokenLearner applies learned spatial attention functions to aggregate intermediate features into a compact set of adaptive tokens~\cite{ryoo2021tokenlearner}.

\subsection{Token merging, pooling, and attention based scores}
PatchMerger inserts a learned module between transformer layers to merge tokens and shorten the sequence~\cite{renggli2022patchmerger}. ToMe progressively combines similar tokens with a lightweight bipartite matching procedure and can be applied to pretrained Vision Transformers without retraining~\cite{bolya2023tome}. Token Pooling approximates a set of intermediate tokens by clustering them to minimize the reconstruction error introduced by downsampling~\cite{marin2023tokenpooling}. Beyond Attentive Tokens combines importance derived from class attention with token diversity through decoupling and merging~\cite{long2023beyond}. WeiToP trains an early pruning module for visual place recognition by distilling aggregation induced token importance~\cite{zeng2026weitop}.

Attention also provides a direct source of token scores. Abnar and Zuidema introduced attention rollout and attention flow as post hoc methods for propagating attention across transformer layers~\cite{abnar2020attention}; we include their rollout construction as one of four attention baselines. Su et al.\ derive Col-Ln from R\'enyi entropy and rank tokens by the $\ell_n$ norm of attention columns~\cite{su2026renyientropy}. Aizawa and Igaue use the entropy of each attention distribution associated with each patch as a pruning criterion and extend the formulation from Shannon to R\'enyi entropy~\cite{aizawa2026renyiattention}.

Most acceleration methods above couple a scoring or adaptation rule with a particular pruning, merging, or pooling mechanism. REDI instead evaluates a supervised, class conditioned scoring signal under a fixed operator. Holding the operator constant separates the comparison of scoring rules from the design of the reducer and makes attention only and corpus aware scores directly comparable at the same token budget.

\subsection{Visual words and corpus weighting}
Classical visual retrieval transferred indexing and term weighting ideas from text retrieval to images. Video Google quantizes local region descriptors into visual words, indexes them with an inverted file, and applies TF-IDF weighting for object retrieval~\cite{sivic2003videogoogle}. The bag of keypoints model represents an image as a histogram of quantized local descriptors for visual categorization~\cite{csurka2004bags}. Vocabulary trees organize visual words hierarchically to support large scale search~\cite{nister2006vocabulary}, while Philbin et al.\ combine large vocabularies with spatial verification for object retrieval~\cite{philbin2007object}. Salton and Buckley review term weighting schemes based on term frequency and collection statistics~\cite{salton1988term}, and Robertson analyzes the theoretical basis of inverse document frequency~\cite{robertson2004idf}.

Recent work has applied visual vocabularies and corpus weighting to learned visual units. The Visual Word Tokenizer compresses ViT inputs using either intra image pixel statistics or an inter image visual vocabulary; its inter image variant averages patches matched to the same visual word~\cite{gee2025vwt}. Wang et al. use TF-IDF to identify class discriminative CNN channels for federated unlearning~\cite{wang2022federated}. Heo et al. use TF-IDF based channel scoring over DINOv2-CLIP aligned features to construct class specific semantic prototypes for few-shot detection~\cite{heo2026prototype}. Cao et al.\ use TF-IDF to identify class related tokens as one component of a broader structural score for assessing and guiding dataset distillation~\cite{cao2026structural}. BM25-V applies BM25 to sparse visual word activations derived from late layer ViT patch features, aggregating patch activations into image level representations for retrieval~\cite{han2026bm25v}. These studies provide relevant precedents for visual word compression and corpus weighting in learned visual spaces. To our knowledge, none of them uses a corpus score indexed by the ground truth class together with image specific attention from the dense path to construct an offline reference ranking for individual patch tokens under a fixed reduction operator.

\section{Method}

\subsection{Problem setting and pipeline}
We use the frozen DINOv3 ViT-B/16 LVD-1689M checkpoint%
\footnote{\texttt{facebook/dinov3-vitb16-pretrain-lvd1689m};
\url{https://huggingface.co/facebook/dinov3-vitb16-pretrain-lvd1689m}}, released for the LVD-1689M setting~\cite{simeoni2025dinov3}. Following the Vision Transformer tokenization scheme~\cite{dosovitskiy2021vit}, the model represents a $224\times224$ input crop as 196 patch tokens on a $14\times14$ grid, one CLS token, and four register tokens, giving a sequence of 201 tokens. We refer to this unreduced configuration as the dense path. The encoder has the standard ViT-B dimensions of 12 transformer blocks, hidden width $D=768$, and 12 attention heads~\cite{dosovitskiy2021vit}. The DINOv3 checkpoint uses axial rotary position embeddings (RoPE)~\cite{simeoni2025dinov3}.

The REDI pipeline comprises two stages: offline computation of a supervised score for patch ranking and reduced evaluation with a fixed token reduction operator. To construct the corpus component, we use the ImageNet-1K training split~\cite{russakovsky2015ilsvrc}.
For each training image, we form a $224 \times 224$ center crop and use all 196 patch representations from the final block to fit a visual vocabulary. The resulting patch assignments are used to construct a TF-IDF table indexed by ImageNet class and visual word. For each validation image, we extract all 196 patch representations from the final block from each of four transformed views and quantize them with the same vocabulary. The ground truth class is then used to retrieve TF-IDF scores for individual patches, which are aligned to the $14 \times 14$ grid of a reference center crop and aggregated into a single map. A separate forward pass on the reference center crop produces an attention map on the same grid. REDI independently normalizes the TF-IDF and
attention maps, multiplies them elementwise, and normalizes the product to obtain the REDI score. The score is precomputed from the ground truth class and dense path attention, so the reduced run is a controlled reference evaluation.

During reduced evaluation, the fixed keep, merge, and compress operator is applied immediately before the first transformer block. Sorting the patch scores assigns 98 patches to the keep set, 49 to the merge set, and 49 to the compress set. Patches in the merge set are aggregated into destination keep tokens, whereas patches in the compress set are aggregated into four compressed tokens using a fixed spatial partition. The CLS token and four register tokens pass through the reduction operator unchanged.
The resulting sequence contains one CLS token, four register tokens, 98 updated keep tokens, and four compressed tokens, for a total of 107 tokens. All 12 frozen transformer blocks process this reduced sequence.

After the final block, the four register embeddings are excluded. The final CLS embedding and the mean of the final patch embeddings are $\ell_2$-normalized separately. The two normalized embeddings are then averaged, and the result is $\ell_2$-normalized to obtain the feature vector supplied to the linear classifier. The patch mean is computed over 196 embeddings in the dense path and 102 embeddings in the reduced path. The classifier is trained once on
these vectors from the dense path and reused unchanged for reduced evaluation. Its optimization settings are specified in the experimental protocol.

\subsection{Supervised visual TF-IDF}

\paragraph{Visual vocabulary.}
Our visual vocabulary follows the bag of visual words abstraction used in image retrieval and categorization~\cite{sivic2003videogoogle,csurka2004bags}.
Unlike those methods, which quantize handcrafted local descriptors, we quantize frozen DINOv3 patch representations. For every ImageNet training image, we resize the shorter side to 256 pixels, take a $224 \times 224$ center crop, and extract all 196 patch representations from the final block before the model's final normalization. After $\ell_2$ normalization, these representations are clustered with spherical $k$-means~\cite{dhillon2001spherical} using $K=512$
centroids. The codebook is initialized deterministically and refined by three full passes over the training feature set, each comprising patch assignment followed by centroid recomputation. Each patch is assigned to the centroid with the highest cosine similarity, and the centroid index is used as its visual word.

\paragraph{Class and corpus statistics.}
Our weighting scheme is inspired by classical TF-IDF term
weighting~\cite{salton1988term,robertson2004idf} and by its application to visual words in Video Google~\cite{sivic2003videogoogle}. The formulation retains image level term frequency and corpus level inverse document frequency. The supervised class contrast, together with the smoothing, clipping, and multi-view aggregation choices, is specific to REDI. For a training image $d$ and visual word $k$, let $n_d(k)$ be the number of its 196 patches assigned to $k$. The term frequency within the image is
\begin{equation}
\mathrm{tf}_{d}(k)=\frac{n_d(k)}{196}.
\label{eq:tf}
\end{equation}
For class $y$, let $\bar p_y(k)$ denote the average of
$\mathrm{tf}_{d}(k)$ over training images in class $y$, and let
$\bar p_{\neg y}(k)$ denote the corresponding average over all remaining training images. Both quantities are floored at $10^{-8}$ before taking logarithms. We define the positive class contrast term as
\begin{equation}
 r_y(k)=
 \max\!\left\{0,
 \log \bar p_y(k)-\log \bar p_{\neg y}(k)
 \right\}\sqrt{\bar p_y(k)}.
\label{eq:classcontrast}
\end{equation}
The positive part retains only visual words that are more prevalent in class $y$ than outside it. The square-root factor weights the contrast by support within the class while remaining sublinear in frequency.

Let $N$ be the number of ImageNet training images, and let
$\mathrm{df}(k)$ be the number of training images containing at least one patch assigned to word $k$. The inverse document frequency is
\begin{equation}
\mathrm{idf}(k)=
\operatorname{clip}_{[0,8]}\!\left(
\log\frac{N+1}{\mathrm{df}(k)+1}
\right).
\label{eq:idf}
\end{equation}
Adding one to the numerator and denominator provides smoothing. Clipping to $[0,8]$ is an implementation choice used consistently in all reported experiments. The resulting table indexed by class is
\begin{equation}
\mathrm{TFIDF}_{\mathrm{gt}}(y,k)
=
 r_y(k)\,\mathrm{idf}(k).
\label{eq:tfidfgt}
\end{equation}
This construction combines corpus level rarity with REDI's supervised class contrast. A large entry requires positive contrast for class $y$, non-negligible frequency within that class, and relative rarity across the training corpus.

\paragraph{Validation TF-IDF map.}
For a vector $q\in\mathbb{R}^{196}$, let
$q_{\min}=\min_j q_j$ and $q_{\max}=\max_j q_j$. We normalize each image map as
\begin{equation}
\mathcal{N}(q)_i=
\begin{cases}
\dfrac{q_i-q_{\min}}
      {q_{\max}-q_{\min}+10^{-8}},
& q_{\max}-q_{\min}>10^{-8},\\[5pt]
0, & \text{otherwise}.
\end{cases}
\label{eq:norm01}
\end{equation}
Thus, a nonconstant map is scaled approximately to $[0,1]$, whereas a constant map is replaced by the zero vector.

For validation view $\nu$, let $z_i^{(\nu)}$ be the visual word assigned to the final block representation at patch position $i$, and let $y$ be the image's ground truth class. The raw patch score and normalized view map are
\begin{equation}
 \tau_i^{(\nu)}
 =\mathrm{TFIDF}_{\mathrm{gt}}\!\left(y,z_i^{(\nu)}\right),
 \qquad
 t^{(\nu)}=\mathcal{N}\!\left(\tau^{(\nu)}\right).
\label{eq:viewtfidf}
\end{equation}
Because the lookup uses the ground truth class $y$, the resulting map is a ground truth conditioned reference score.

Our validation procedure uses four views. After resizing the shorter side to 256 pixels, we take a $224 \times 224$ center crop, its horizontal flip, and two deterministic random resized crops. The random crops use a crop area scale of $[0.60,1.00]$ and an aspect ratio of $[3/4,4/3]$. Crop parameters and horizontal flips are deterministically seeded by the image index; each random crop is flipped with probability $0.5$. Every view is represented at $224 \times 224$ before feature extraction. 

The four score maps are aligned to the grid of the reference center crop. A flipped map is first restored to its original orientation. The reference patch centers are then expressed in the coordinate system of each transformed crop, and bilinear sampling maps the scores back to the $14 \times 14$ grid. A validity mask excludes reference locations outside a transformed crop. For
reference patch $i$, let $V_i$ be the set of views that cover that location, and let $\widetilde t_i^{(\nu)}$ be the corresponding remapped score. We compute
\begin{equation}
\begin{aligned}
 \mu_i
 &=\frac{1}{|V_i|}
   \sum_{\nu\in V_i}\widetilde t_i^{(\nu)},
 &
 \xi_i
 &=\max_{\nu\in V_i}\widetilde t_i^{(\nu)},\\
 t
 &=\mathcal{N}\!\left(0.8\,\mu+0.2\,\xi\right).
\end{aligned}
\label{eq:viewaggregate}
\end{equation}
The center crop covers every reference patch location, so $V_i$ is never empty. We fix the aggregation coefficients to $0.8$ and $0.2$ in all experiments.

\subsection{Attention scores}
Attention scores are computed in a separate forward pass on the reference center crop. In the self-attention matrix, rows correspond to query positions and columns correspond to key positions~\cite{guo2023robustifying}. We aggregate the column associated with each patch over all query positions and attention heads. We use the descriptive term \emph{incoming attention mass} for this accumulated column weight; related ViT analyses study how attention behavior varies across supervision regimes~\cite{walmer2023teaching}. This final block statistic serves as the primary image conditioned ranking signal. Let $A^{(L,h)}\in\mathbb{R}^{T\times T}$ denote the attention matrix after softmax for head $h$ in block $L=12$, with $H=12$ heads and $T=201$ tokens. For patch position $i$, let $j(i)$ denote its key column after the CLS token and four register tokens. The raw score is
\begin{equation}
\widetilde a_i
=
\frac{1}{HT}
\sum_{h=1}^{H}
\sum_{q=1}^{T}
A^{(L,h)}_{q,j(i)}.
\label{eq:attentionmassraw}
\end{equation}
This quantity is the mean of patch $i$'s key column over all 201 query positions and 12 heads. This statistic is image conditioned but remains dense path information, since it is extracted before applying the reduced operator. The primary attention map is
$s_{\mathrm{mass}}=\mathcal{N}(\widetilde a)$.

We also evaluate three alternatives. The CLS mean map averages, across heads, attention from the CLS token to the patch tokens in the final block. The CLS maximum map instead takes the maximum across heads. Our rollout baseline follows the layerwise matrix composition of Abnar and Zuidema~\cite{abnar2020attention}, adapted to the DINOv3 token sequence. It averages heads within each block, adds the identity matrix, normalizes each row, and composes the resulting transition matrices across all 12 blocks. The rollout score is the final CLS row restricted to the 196 patch
columns. The exact tensor reductions are given in the supplementary material. Each attention map is combined with the corpus term to produce the score used for patch ranking and aggregation.

\subsection{REDI score}

Let $t\in\mathbb{R}^{196}$ be the supervised TF-IDF map and let
$a\in\mathbb{R}^{196}$ be an attention map. The primary REDI variant uses $a=s_{\mathrm{mass}}$. REDI normalizes both inputs, multiplies them elementwise, and normalizes the product:
\begin{equation}
\widehat t=\mathcal{N}(t),
\qquad
\widehat a=\mathcal{N}(a),
\qquad
s=\mathcal{N}\!\left(\widehat t\odot\widehat a\right),
\label{eq:hybrid}
\end{equation}
where $\odot$ denotes elementwise multiplication. The normalized product defines the REDI score. A high REDI score therefore requires both a high class conditioned corpus weight and a high value under the selected attention signal. The same rule is applied to each alternative attention map. The REDI score has two roles in the fixed operator: its rank assigns patches to the keep, merge, and compress sets, and its magnitude weights patch contributions during merging and compression. Merge destinations are selected separately using feature similarity and spatial proximity, whereas compress cell membership depends only on grid location. Section~\ref{sec:operator} specifies these operations.

\subsection{Fixed keep, merge, and compress operator}
\label{sec:operator}

We fix the operator to compare scoring signals under a common reduction mechanism. It is conceptually related to ViT acceleration methods that preserve attentive tokens while fusing inattentive ones (EViT)~\cite{liang2022evit}, progressively merge similar tokens (ToMe)~\cite{bolya2023tome}, or approximate
a token set with a smaller set by minimizing downsampling reconstruction error (Token Pooling)~\cite{marin2023tokenpooling}. Its design draws on these general principles, but the fixed budget, role assignment, spatial bias, and aggregation weights are specific to our evaluation protocol rather than an implementation of any single prior method.

\paragraph{Role assignment.}
For any scoring configuration, let $s_i$ denote the normalized score assigned to patch $i$. We sort the 196 patch scores in descending order. The highest $n_R=\operatorname{round}(0.50\times196)=98$ scores define the keep set $R$, the next $n_M=\operatorname{round}(0.25\times196)=49$ define the merge set $M$, and the remaining $n_U=196-n_R-n_M=49$ define the compress set $U$. The three sets are disjoint and together contain all 196 patches. The rank of $s_i$ determines the role of each patch, while its magnitude is subsequently used to weight aggregation. The scores are precomputed along the dense path, whereas token reduction is applied to the patch embeddings immediately before the first transformer block. Let $x_i\in\mathbb{R}^{D}$ denote the embedding of patch $i$ at this point, and let $g_i\in\{0,\ldots,13\}^2$ denote its coordinate on the $14\times14$ patch grid.

\paragraph{Merge destination selection.}
Each patch $i\in M$ is assigned to one patch in the keep set. For $i\in M$, 
\begin{equation}
\begin{aligned}
 \rho_{ir}
 &=1-\frac{\lVert g_i-g_r\rVert_2}{13\sqrt{2}},\\
 m(i)
 &=\arg\max_{r\in R}
 \left\{\cos(x_i,x_r)+0.30\,\rho_{ir}\right\}.
\end{aligned}
\label{eq:mergeassignment}
\end{equation}
Here $\rho_{ir}\in[0,1]$ denotes normalized spatial proximity, and
$13\sqrt{2}$ is the maximum Euclidean distance on the $14\times14$ grid. Once the role sets have been determined, the merge destination depends only on cosine similarity between the pre-block patch embeddings and on spatial proximity. The patch score $s_i$ is not part of the destination criterion. The spatial coefficient is fixed at $0.30$ and is not learned.

\paragraph{Score weighted merge update.}
Once all merge destinations have been assigned, the normalized patch scores are used to weight aggregation. For each $i\in M\cup U$, we define $\omega_i=\max\{s_i,0\}+10^{-4}$, where the small positive offset prevents a zero aggregation weight. Each keep token is assigned a self weight of one.
The updated embedding of $r\in R$ is
\begin{equation}
\widetilde x_r=
\frac{
x_r+\sum_{i\in M:\,m(i)=r}\omega_i x_i
}{
1+\sum_{i\in M:\,m(i)=r}\omega_i
},
\qquad r\in R.
\label{eq:mergeupdate}
\end{equation}
If no patch in $M$ is assigned to $r$, then $\widetilde x_r=x_r$.
Equation~\eqref{eq:mergeassignment} determines the destination of each merge patch, whereas Eq.~\eqref{eq:mergeupdate} determines its contribution to the updated keep token.

\paragraph{Spatial compression.}
The compress set is summarized separately rather than assigned to keep tokens. We divide the $14\times14$ patch grid into four non-overlapping $7\times7$ cells. For each cell $c$, let $U_c\subseteq U$ denote the compress set patches located within that cell. If $U_c$ is nonempty, its patches are summarized by one score weighted token,
\begin{equation}
u_c=
\frac{\sum_{i\in U_c}\omega_i x_i}
{\sum_{i\in U_c}\omega_i}.
\label{eq:compressupdate}
\end{equation}
If $U_c$ is empty, the same weighted average is computed over the full compress set $U$, ensuring that the operator always produces four summary tokens. A summary token from a nonempty cell inherits the RoPE position of the highest scoring patch in $U_c$. For an empty cell, it inherits the position of the highest scoring patch in $U$. This construction preserves an existing patch position rather than introducing a synthetic coordinate.

The updated tokens in $R$ are restored to their original spatial order and retain their original RoPE positions. The four compressed tokens are then appended. Together with the unchanged CLS token and four register tokens, these outputs form the 107-token sequence processed by the frozen encoder.

\subsection{Analytical cost estimate}

The sequence length decreases from 201 to 107 tokens, a reduction of 46.8\%. We estimate the resulting reduction in the dominant matrix multiplications of a standard ViT-B block. The block contains four $D\times D$ attention projections and an MLP with two layers and hidden width $4D$~\cite{dosovitskiy2021vit}. This structure gives
\begin{equation}
C_{\mathrm{block}}(T,D)
=
12TD^2+2T^2D.
\label{eq:blockcompute}
\end{equation}
The linear term comprises $4TD^2$ for the query, key, value, and output projections and $8TD^2$ for the two MLP projections. The quadratic term accounts for the product of queries and keys and the multiplication of attention weights by values.

With $D=768$, $C_{\mathrm{block}}(107,768)/C_{\mathrm{block}}(201,768)=52.2\%$. Thus, the estimated arithmetic cost of these matrix multiplications is reduced by 47.8\% in each transformer block. Because reduction occurs before block 1, the same ratio applies to all 12 transformer blocks. The quadratic attention term retains $(107/201)^2=28.3\%$ of its value in the dense path, corresponding to a reduction of 71.7\%.

These estimates report the change in the dominant encoder matrix multiplications caused by the shorter token sequence. We discuss the scope of these estimates in Sec.~\ref{sec:limitations}.

\section{Experimental results}
\label{sec:results}
\subsection{Evaluation protocol}
We use the ImageNet-1K training split~\cite{russakovsky2015ilsvrc} to fit the visual vocabulary, estimate class conditioned TF-IDF statistics, and train a single linear classifier. All accuracies are reported on the complete validation split of 50,000 images. The pretrained DINOv3 ViT-B/16 backbone remains frozen. The classifier is trained for 20 epochs on features extracted from dense passes using AdamW, a batch size of 4096, an initial learning rate of $3\!\times\!10^{-3}$ with cosine decay, weight decay of $10^{-2}$, and no label smoothing. Feature extraction, classifier training, and evaluation use FP32 without automatic mixed precision. Cached score maps are stored in FP16 and converted to FP32 when loaded.

Every reduced configuration uses the same frozen backbone, classifier, 107 token budget, merge and compression rules, and RoPE handling defined in Sec.~\ref{sec:operator}. Across reduced evaluations, only the patch scoring signal used for ranking and aggregation changes. Since REDI scores are precomputed, the experiments compare reference rankings under a shared reducer.

\subsection{Main comparison}
\begin{table}[t]
\centering
\caption{ImageNet-1K accuracy for dense and reduced evaluation. The REDI score is the normalized product of class-conditioned visual TF-IDF and incoming attention mass. All reduced configurations use the same 107-token budget, frozen backbone, classifier, and reduction operator.}
\label{tab:main}
\resizebox{\columnwidth}{!}{%
\begin{tabular}{lrrrr}
\toprule
Configuration & Tokens & Top-1 & $\Delta$ dense & Top-5\\
\midrule
Dense baseline & 201 & 83.514 & 0.000 & 96.902\\
Supervised $\mathrm{TFIDF}_{\mathrm{gt}}$ & 107 & 81.796 & -1.718 & 96.236\\
Incoming attention mass & 107 & 82.634 & -0.880 & 96.454\\
\midrule
\textbf{\textsc{REDI} with incoming attention mass} & \textbf{107} & \textbf{84.706} & \textbf{+1.192} & \textbf{97.442}\\
\bottomrule
\end{tabular}}
\end{table}

Table~\ref{tab:main} compares the dense path with reduced evaluation under the fixed operator. The dense path reaches 83.514\% Top-1 with 201 tokens. At 107 tokens, $\mathrm{TFIDF}_{\mathrm{gt}}$ alone reaches 81.796\% Top-1 and incoming attention mass alone reaches 82.634\%. Their normalized product reaches 84.706\% Top-1, giving gains of 1.192 points over dense, 2.910 over $\mathrm{TFIDF}_{\mathrm{gt}}$, and 2.072 over incoming attention mass under the same reducer.

\subsection{Consistency across attention formulations}
\begin{table}[t]
\centering
\caption{Paired comparisons at the 107-token budget. ``Gain'' denotes the Top-1 difference between the REDI product formed with the listed attention score and its attention-only counterpart.}
\label{tab:attentionpartners}
\resizebox{\columnwidth}{!}{%
\begin{tabular}{lrrrr}
\toprule
Attention score & Attention Top-1 & REDI Top-1 & Gain & REDI Top-5\\
\midrule
Incoming attention mass & 82.634 & \textbf{84.706} & +2.072 & 97.442\\
Attention rollout & 81.314 & 84.644 & +3.330 & \textbf{97.446}\\
CLS attention max & \textbf{82.834} & 84.574 & +1.740 & 97.386\\
CLS attention mean & 82.788 & 84.524 & +1.736 & 97.384\\
\bottomrule
\end{tabular}}
\end{table}

Table~\ref{tab:attentionpartners} pairs each attention only configuration with the REDI product formed from the same attention score. Combining each attention score with $\mathrm{TFIDF}_{\mathrm{gt}}$ improves Top-1 accuracy by 2.072 points for incoming attention mass, 3.330 points for attention rollout, 1.740 points for CLS attention maximum, and 1.736 points for CLS attention mean. Every attention only configuration remains below the 83.514\% dense baseline, whereas every REDI product exceeds it by 1.010--1.192 points.

\subsection{Sequence and analytical compute reduction}
\begin{table}[t]
\centering
\small
\setlength{\tabcolsep}{3.0pt}
\resizebox{\columnwidth}{!}{%
\begin{tabular}{lccc}
\toprule
Quantity & Dense & Reduced path & Reduction \\
\midrule
Individually represented patches & 196 & 98 & 50.0\% \\
Patch representations after reduction & 196 & 102 & 48.0\% \\
Total sequence tokens & 201 & 107 & 46.8\% \\
Quadratic attention term & $201^2$ & $107^2$ & 71.7\% \\
Estimated dominant block arithmetic & 1.000 & 0.522 & 47.8\% \\
\bottomrule
\end{tabular}}
\caption{Analytical sequence and compute reductions at the block level after the fixed operator forms the 107-token sequence. The estimates exclude score construction, reduction overhead, and hardware-dependent effects.}
\label{tab:compute}
\end{table}

At 107 tokens, the estimated dominant matrix-multiplication cost per frozen block and the quadratic attention term fall by 47.8\% and 71.7\%, respectively. The scope of these analytical estimates is discussed in Sec.~\ref{sec:limitations}.

\section{Discussion}
Combining the same class conditioned corpus term with attention improves reduced classification across all four attention formulations under an otherwise identical setup. Because the backbone, classifier, token budget, and reduction operator remain fixed, this repeated pattern indicates that the corpus term contributes ranking information that is not captured by any evaluated attention score alone.

The two components encode different forms of context. Attention reflects image specific token interactions but contains no explicit class or corpus level statistics. The TF-IDF term contributes class contrast, within class support, and corpus rarity, but assigns the same lookup value to patches mapped to the same visual word within a view. Their product therefore favors patches supported by both class conditioned corpus statistics and image conditioned attention. The gains obtained with all four attention formulations suggest that this complementarity is not specific to incoming attention mass.

Supplementary DeiT~III~\cite{touvron2022deit3} and MAE~\cite{he2022mae} diagnostics show that the hybrid signal remains close to dense accuracy across the two added backbones, but those experiments use a different classifier protocol and are therefore treated as diagnostic.

The score affects the reduced representation in two ways. Its rank determines which patches remain explicit, which are merged, and which contribute to spatial summaries; its magnitude controls the contribution of merged and compressed patches during aggregation. The classifier is trained only on dense features and reused without adaptation, so differences in reduced accuracy measure how well each reduced representation supports the same linear decision rule. Under this protocol, the joint score yields higher accuracy than either component alone.

The gain over the dense baseline is specific to the present backbone, classifier, token budget, and reducer. The fixed reducer is useful because it isolates the contribution of corpus level class information without implying a general advantage of shorter sequences or optimality of the operator.

\section{Limitations and future work}
\label{sec:limitations}
The main evaluation uses DINOv3 ViT-B/16; supplementary DeiT~III and MAE experiments use model native classifiers and a separate classifier adaptation protocol. REDI also relies on the ground truth class and dense path attention, so it should be interpreted as a supervised reference for patch relevance under a fixed reducer rather than a deployable acceleration pipeline. Therefore, the accuracy gain over the dense baseline should not be read as evidence that shorter sequences are generally superior, but as evidence that this offline reference ranking can allocate the fixed token budget more favorably for the reused classifier.

The reported sequence length and arithmetic reductions describe the reduced encoder path after REDI scores are available. They do not include score construction, dense attention extraction, reduction overhead, operations outside the dominant matrix multiplications, memory traffic, or hardware effects. Future work should replace the ground truth class lookup and dense attention map with predicted or early layer approximations, and test broader architectures, tasks, budgets, and reducer designs.

\section{Conclusion}
REDI ranks DINOv3 patch tokens by combining class indexed visual-word statistics with image specific attention under a fixed reduction budget. Across four attention formulations, the combined score improves reduced classification over either component alone while using the same \textit{keep}, \textit{merge}, and \textit{compress} operator. The results show that corpus level class statistics provide a useful complementary signal for controlled token allocation and motivate future work on deployable approximations.

{
    \small
    \bibliographystyle{ieeenat_fullname}
    \bibliography{main}
}

\end{document}